\documentclass[twoside,11pt]{article}
\usepackage{hyperref}
\usepackage{natbib}
\usepackage{jair, rawfonts}
\usepackage{booktabs}
\usepackage{multirow}
\usepackage{rotating}
\usepackage{amsmath}
\usepackage{amssymb}
\usepackage[utf8]{inputenc}

\usepackage{tablefootnote}
\usepackage{adjustbox}

\renewcommand{\cite}[1]{%
  \citep{#1}%
}

\newcommand{\citeTable}[1]{%
  \citet{#1}%
}

\begin{document}

\ShortHeadings{A Survey of Language-Based Communication in Robotics}
{Hunt et al.}
\firstpageno{99}

\title{A Survey of Language-Based Communication in Robotics}

\author{\name William Hunt \email W.Hunt@soton.ac.uk \\
       \name Sarvapali D. Ramchurn \email sdr1@soton.ac.uk \\
       \name Mohammad D. Soorati \email M.Soorati@soton.ac.uk \\
       \addr University of Southampton\\
       University Rd, Southampton SO17 1BJ
       }

\maketitle

\begin{abstract}
    Embodied robots which can interact with their environment and neighbours are increasingly being used as a test case to develop Artificial Intelligence. This creates a need for multimodal robot controllers that can operate across different types of information, including text. Large Language Models are able to process and generate textual as well as audiovisual data and, more recently, robot actions. Language Models are increasingly being applied to robotic systems; these Language-Based robots leverage the power of language models in a variety of ways. Additionally, the use of language opens up multiple forms of information exchange between members of a human-robot team. This survey motivates the use of language models in robotics, and then delineates works based on the part of the overall control flow in which language is incorporated. Language can be used by a human to task a robot, by a robot to inform a human, between robots as a human-like communication medium, and internally for a robot's planning and control. Applications of language-based robots are explored, and numerous limitations and challenges are discussed to provide a summary of the development needed for the future of language-based robotics.\footnote{Links to each paper and, if available, source code are provided at the accompanying webpage: \href{https://sooratilab.com/publications/papers/2024/A-Survey-of-Language-BasedCommunication-in-Robotics.php}{https://sooratilab.com/publications/2024/A-Survey-of-Language-BasedCommunication-in-Robotics.php}}. 
\end{abstract}

\maketitle

%%%%%%%%%%%%%%%%%%%%%%%%%%%%%%%%%%%%%%%%%%%%%%%%%%%%%%%%%%%%%%%%%%%%%%%%%%%%%%%%
\section{Introduction}
A popular trend in Artificial Intelligence is toward powerful, multimodal models that operate on multiple domains of input and output~\cite{reed2022generalist}. This often centres around foundational models based on vision and language~\cite{di2023towards}. This is especially true in the field of robotics, where future robotic systems could operate on a single architecture that combines learning, understanding, and actions in different modalities~\cite{driess2023palm}. Core to this approach is the integration of language models; not only does this potentially improve the power and potential of robotic systems, but it also introduces many promising interaction opportunities. Language can be included in various elements of the total architecture, each providing different advantages. 

\begin{figure}[h!]
    \centering
    \includegraphics[width=\linewidth]{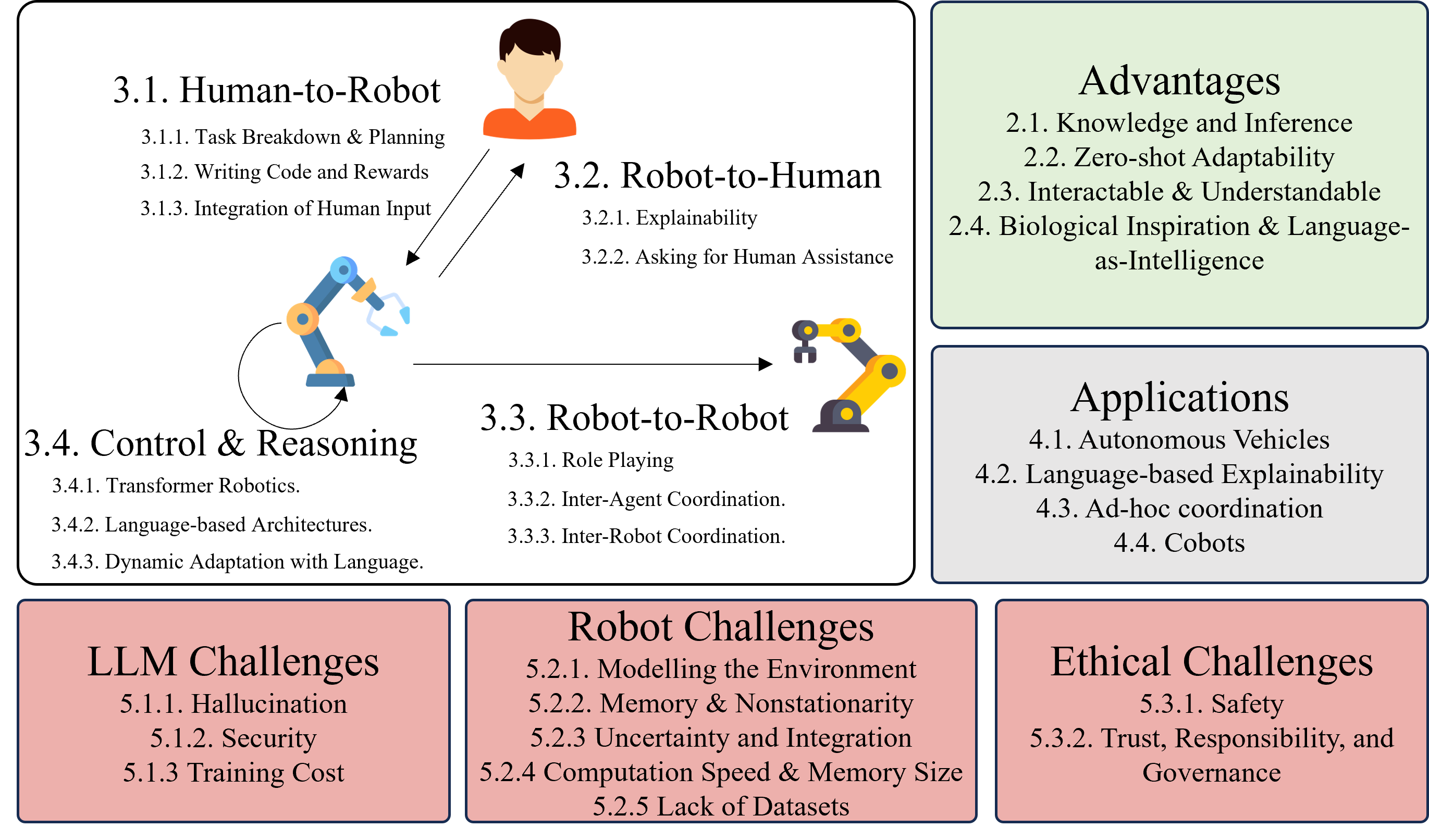}
    \caption{Overview of this survey. Works are categorised by direction of communication between members of the human-robot team. Advantages, applications, and limitations are also discussed.}
    \label{fig:overview}
\end{figure}

Large Language Models (LLMs) are a rapidly advancing technology that shows promise in many diverse areas of science and engineering~\cite{yang2023harnessing}. Among the many potential advantages of a language-based AI landscape is that natural language encapsulates deep meaning and complex relationships between semantic concepts; a sentence of only a few words may represent a detailed idea, and minor alterations can substantially alter the meaning~\cite{crocker2016information}. Virtually all human interaction, coordination, and problem-solving is in some way facilitated by language, and yet robots have traditionally never benefited from this. Despite being intended to replace human workers, robots in the past have generally been controlled through direct teleoperation, predefined paths, or reinforcement learning~\cite{kober2013reinforcement}. An emerging trend in robotics is the integration of LLMs into the robotic command and control structure~\cite{zeng2023large}. This development goes beyond the simple paradigm of labelling predefined, hand-written behaviours which can be commanded by a human user, and instead treats language as an input space like any other. Language models instead utilise the concept of ``embeddings'' to capture the semantic meaning of a piece of text~\cite{reimers2019sentence}. This allows them to operate more freely and flexibly with respect to whatever wording or structure of textual input is provided. It also presents an opportunity for greater generalisation, where the language model can learn from examples in a manner that does not need to conform to a given structure.

\citeTable{kim2024survey} explored the use of LLMs in robots by dividing them based on the application of the LLM; grouping them into communication, planning, perception, and control methods. Their study of communication covered language generation and understanding including some human-interaction aspects. This survey offers a review of the various parts of a robot or robots on which language can be applied. \citeTable{wang2024survey} focus only on the single-agent case, they explore LLMs for agent memory as well as other parts of the architecture of an agent into which language can be integrated. \citeTable{guo2024large} reviewed LLMs for multi-agent systems, exploring different applications of the technology and the various architecture designs. Each of these works covers the agent-based model of LLMs very well, although both are outside the scope of robotics. \citeTable{zeng2023large}, like \citeTable{kim2024survey}, divided works by application in robotics into the areas of perception, decision-making, control, and interaction. They also include a glossary of terms and models that summarises the key ideas and models for LLM-based robots. \citeTable{zhang2023large} provided an exploration of human-robot interaction by categorising the use of LLMs for physical, teleoperation, and conversation methods. \citeTable{atuhurra2024large} covered multiple applications and architectures for LLMs as well as societal norms surrounding them. Both of these works are highly relevant to the relationship between a language-based robot and its operators. 

None of the previous surveys explore the entire context of a mixed team of humans and robots. A scenario where one or more humans must control or operate alongside a team of multiple agents demands an understanding of how language could be employed in every part of the system. Viewing the team of robots and operators as a collective who communicate, report, and control each other and themselves through the use of language provides an appropriate grouping of works based on where in the system they could feasibly be deployed. This survey aims to fill this gap by classifying and exploring works based on the part of the overall team structure in which language can be integrated

Language presents several opportunities in robotics, both from a research perspective and one of the future systems deployed in the real world (section \ref{sec:whyLang}). We propose that any system with one or many robots and one or many humans working alongside each other can be considered in a manner where language can be inserted as a component of any interface between them (section \ref{sec:paradigm}). Language can now be used as an input space for robotic models so they can respond to hand-written instructions from humans without the need to conform to a rigid format or grammar (section \ref{sec:H2R}). Robots can also generate language that is used to explain or validate their decisions and behaviour to a human (section \ref{sec:R2H}). This can be further extended to have robots communicate with each other using language as both an input and output, simulating human collectives who plan and coordinate in this way (section \ref{sec:R2R}). Emerging research integrates language internally into the model as part of its architecture (section \ref{sec:Ri}). We describe these in detail and then summarise some potential future applications of this concept (section \ref{sec:App}), as well some key limitations and challenges in this area (section \ref{sec:Lim}).

%%%% Made a pdf version but side-by-side the png is better
%\begin{figure}[h!]
%    \centering
%    \includegraphics[trim={3.9cm 5.8cm 5.5cm 3.1cm},clip, width=\linewidth]{Fig1Only.pdf}
%    \caption{Overview of this survey. Works are %categorised by direction of communication, and advantages, applications, and limitations are discussed.}
%    \label{fig:overview}
%\end{figure}

\section{Why Use Language?}\label{sec:whyLang}
The scope of research using LLMs is reasonably broad. There are being applied to a wide variety of domains outside of standard text generation including education~\cite{kasneci2023chatgpt}, medicine~\cite{thirunavukarasu2023large,lievin2024can}, programming~\cite{xu2022systematic}, and video games~\cite{AbiRaad2024Scaling}. They are also beginning to be employed in multi-agent systems~\cite{guo2024large,wang2024survey}, human-robot interaction~\cite{zhang2023large,atuhurra2024large} and robotics generally~\cite{kim2024survey,zeng2023large}. This section suggests four core reasons to investigate LLMs in robotics.

\subsection{Knowledge and Inference}
Language models are trained on a large and varied corpus of written text, online web data, and community-maintained repositories of information~\cite{minaee2024large,liu2024datasets}. As such, this knowledge is distilled into the models, granting them an extremely broad and often quite deep ‘understanding’~\cite{Hinton_2023}. While the true nature of their ability to reason is debated~\cite{cohen_2023,browning2023language,haggstrom2023large}, they seem capable of combining knowledge and ideas together to produce new outputs~\cite{romera2024mathematical}. This makes them well suited to interpreting the true meaning behind the tasks they are set, as well as being able to expand on observations and details by combining learned knowledge in order to achieve their goals.

\subsection{Zero-shot Adaptability}
Language models have been proposed as excellent few-shot or even zero-shot learners~\cite{brown2020language}. In robotics the environment is unpredictable and unforeseen elements are common. By applying language models to the problem, we can benefit from their ability to guess about the impacts of novel events. For example, a robot may never have trained for a particular terrain or weather, but language can offer an estimation of its impact~\cite{shek2023lancar}. Similarly, new tasks that have not been trained for may be decomposed by a language model into performable substeps, or else the model may prove effective at understanding which parts of a complex task are important to consider and how to handle them. The ability for language models to perform in-context learning may be a useful tool in empowering other approaches, even if not to replace them entirely~\cite{dasgupta2023collaborating}.

\subsection{Interactable and Understandable}
Language is the fundamental means of direct interaction between humans. Both structured and unstructured collectives generally utilise language to coordinate, plan, and exchange information~\cite{mirsky2020penny}. To understand and communicate often requires a dialogical component in order to ground information and establish a shared knowledge and perception of the world~\cite{clark1991grounding}. As an extremely dense embedding of information, a few words can encapsulate deep meaning. The specific framing and interplay between the words allows a listener to comprehend a complex, high-level idea from a simple sentence~\cite{minsky1988societyGrammar}. Allowing robots to use language opens up their command, control, and even reasoning to human intervention; there is no requirement for operators and bystanders to understand programming languages or technical commands. By building robots which can emulate human communication, they can be commanded by humans or can explain their actions in the same way a human worker might. They can also cooperate with each other in an understandable way~\cite{stone2010ad}. A language-based architecture, or one where language is used to interface between parts of a policy or agents in a team allows robots to be swapped-out for humans at any level and also allows humans to understand and assist robots easily. This also makes it much easier for subject matter experts to work alongside robots without technical knowledge of their processes.

\subsection{Biological Inspiration and Language-as-Intelligence}
Much of the development of Artificial Intelligence has been ‘bottom-up’ –- taking inspiration from biological intelligence and building computer systems which model the same structures and processes in the digital realm~\cite{frenkel2021bottom}. Neural networks are a common example of this, where relatively naïve approaches allow us to emulate the way humans learn; as nature has arrived at a formula for human intelligence, this surely must be one means of achieving it digitally, even if not necessarily the best~\cite{Hinton_2023}. Rodney Brooks postulated that embodiment was key to intelligence, making robots a good test case for building AI~\cite{brooks1991intelligence}, while others also argue that language is in some sense core to human thinking and reasoning~\cite{goldstein1977artificial,bloom2001thinking,gleitman2013relations,pinker2003language}, even perhaps that language is a ``system of thought'' \textit{sui generis}~\cite{chomsky2011language}, or at least shapes the way we perceive and experience the world around us~\cite{boroditsky2011language}. It may be the case that empowering physical agents with the ability to interpret, internally reason with, and output language could substantively improve their intelligence. Regardless, language as a planning and communication tool naturally mimics human intelligence and behaviour in many settings, so it may be productive to explore in the context of robots as human-like workers.

\section{The Paradigm for Communication in Multi-Robot Systems}\label{sec:paradigm}
As the number of robots and other intelligent systems increases in prevalence in people's everyday lives, it is prudent to expect a complex combination of such systems interacting. A household environment, for example, may already have multiple robot hoovers, along with networked smart appliances with the facility to effect the environment. The term ``robot'' is hard to clearly define, but we draw on a wide literature which includes a variety of autonomous systems; the overarching movement towards large numbers of interacting platforms makes it important to consider the heterogeneity of robots, and we also consider LLMs employed in agent-based configurations that would naturally extend to robotics and are likely to in the near future.

To motivate this survey, we describe a scenario where one or multiple humans may be interacting in an environment along with one or multiple robots. Humans of course interact, plan, and coordinate using natural language; robots can use natural language wherever possible to interact with humans, and may even use language for some or all of their coordination as well as internal processing. Such a scenario might be a family household with several personal robots, a busy building site with humans and robots working together, or an office with humans and robots moving around it. The remainder of this work will break down each of these elements of a language-based, human-in-the-loop multi-robot system and explore the role of language at each step.

\subsection{Human to Robot Communication}\label{sec:H2R}
The most suited use-case of language models in robotics is the direct commanding of robots. Classical approaches have defined a library of skills and commands with textual labels that can be used to control a robot~\cite{abioye2018multimodal}. However, the introduction of LLMs redefines this paradigm; what was once a discrete input space defined by a programmer can now be learned at a higher level. By considering a textual (or, by extension, visual or gestural) command as encoding some meaning which exists in an embedding space, it can be theorised that a model can interpret an instruction never before seen and internally relate it to previous inputs. This is encapsulated by the fundamental concept of machine learning --- generalisation, but it is an important and nontrivial step to move from generalising based on training to data, to understanding and acting based on the semantic meaning of a sentence. The ``Human to Robot Communication'' category describes work which focuses on using human-style commands that may otherwise be issued to other humans, such as ``pick up the red ball'' and creating models that map this to an action.

\begin{table}
\centering
\caption{A selection of works using language for Human-to-Robot tasking. The ``LLM'' column indicates the most recent LLM they used (not all if they used several variants). \textit{Italics} indicate a simulator, and ``Application'' is left high level to allow papers to be further subgrouped.}
\label{tab:LetRevH2R}
\begin{adjustbox}{max width=\textwidth}
\begin{tabular}{@{}cllll@{}} % Adjust column specifiers (l, c, r) as needed
\toprule
Modality & Paper & LLM & Physical Robot / \textit{Simulation} & Application \\ % Table headers
\midrule
\multirow{9}{*}{\rotatebox{90}{Task Breakdown}}
& \citeTable{dorbala2023can} & GPT-3 & \textit{RoboTHOR} & Generating plans \\
& \citeTable{brohan2023can} & GPT-3.5 & Franka Robot & Generating plans \\ 
& \citeTable{wu2023embodied} &  TaPA & \textit{AI2THOR} & Generating plans \\ 
& \citeTable{kakodkar2023cartier} & GPT-3 & Clearpath Jackal UGV & Generating plans\\ 
& \citeTable{huang2022language} & GPT-3 & \textit{VirtualHome} & Generating plans \\ 
& \citeTable{rana2023sayplan} & GPT-4 & Robot arm on wheels & Generating plans \\ 
& \citeTable{kapelyukh2023dall} & DALL-E & Robot arm & Generating target state \\ 
& \citeTable{wang2023gensim} & GPT-4 & \textit{Ravens} & Generating simulation data \\
& \citeTable{zhi2024pragmatic} & GPT-3 & \textit{VirtualHome} & Determining human intentions \\
\midrule
\multirow{13}{*}{\rotatebox{90}{Code and Rewards}} 
& \citeTable{singh2023progprompt} & GPT-3 %And others
& \textit{Virtual Home} \& Franka Panda & Writing code\\
& \citeTable{hu2024deploying} & Various & \textit{RoboEval} & Writing code\\ 
& \citeTable{lin2023gesture} & GPT-3.5 & Franka Panda & Writing code \\ 
& \citeTable{huang2023visual} & GPT-3 & \textit{AI2THOR} & Writing code \\
& \citeTable{liang2023code} & GPT-3 & UR5e robot arm & Writing code \\ 
& \citeTable{lin2023text2motion} & GPT-3.5 & Franka Panda & Writing code \\
%& \citeTable{lykov2023llm} & GPT-3 & N/A & Write behaviour trees \\ 
& \citeTable{yu2023language} & GPT-4 & \textit{MuJuCo} & Generating reward function \\ 
& \citeTable{hu2023language} & GPT-3.5 & N/A & Generating reward function \\
& \citeTable{zhang2023planning} & GPT-2 & N/A & Writing code \\
& \citeTable{ma2023eureka} & GPT-4 & \textit{IsaacGym} & Writing code \\
& \citeTable{narin2024evolutionary} & GPT-4V & \textit{IsaacGym} & Writing code \\
& \citeTable{yu2023scaling} & GPT-3 & Robot arm & Generating training examples \\
& \citeTable{wu2023tidybot} & PaLM 540B & Kinova Gen3 7-DoF & Writing code \\
\midrule
\multirow{16}{*}{\rotatebox{90}{Integrated Input}} 
& \citeTable{wen2024object} & GPT-3.5 & Franka Robot & Augmenting into model \\
& \citeTable{shah2023lm} & GPT-3 & Clearpath Jackal UGV & Augmenting into input space \\
& \citeTable{ha2023scaling} & GPT-3 \& LLAMA2 & \textit{MuJoCo} & Writing success labelling function \\
& \citeTable{zhu2024incoro} & GPT-3.5 & N/A & Integrating into model \\
& \citeTable{barmann2023incremental} & GPT-3.5 & ARMAR-6 & Integrating as memory \\
& \citeTable{liu2023llm} & GPT-2 & \textit{Overcooked} & Integrating into model \\
& \citeTable{bing2023meta} & GloVe & \textit{Metaworld} and robot arm & Integrating into MDP \\
& \citeTable{ni2023grid} & INSTRUCTOR & PUDUbot2 & Integrating into model \\
& \citeTable{yow2024extract} & S-BERT & xArm-6 manipulator & Integrating into model \\
& \citeTable{bucker2023latte} & BERT & Panda robot arm & Embedding input into architecture\\
& \citeTable{lynch2023interactive} & LCBC\tablefootnote{Language conditioned behavioral cloning} & \textit{Language-Table} and xArm6 & Integrating into model\\
& \citeTable{cui2023no} & GPT-3 & Franka Emika Panda & Modifying with embedding \\
& \citeTable{huang2023voxposer} & GPT-4 & Franka Emika Panda & Modifying with embedding \\
& \citeTable{karamcheti2022lila} & Distil-RoBERTa & Franka Panda & Disambiguating manual input \\
& \citeTable{gu2023rt} & RT-Trajectory & Everyday Robots arm & Generating trajectories \\
& \citeTable{shi2024yell} & GPT-4V & ALOHA & Modifying behaviour \\
\bottomrule
\end{tabular}
\end{adjustbox}
\end{table}

\subsubsection{Task Breakdown and Planning}
The first and most straightforward strategy is for the human to define a high-level mission objective which is broken down by the model into subtasks. This approach can generate plans which directly inform robot actions, although typically via a proxy such as labelled action policies. An early example of this approach is SayCan, which uses an LLM to suggest a suitable action in response to a textual prompt, and then selects the best actions from a list of pretrained affordances~\cite{brohan2023can}. Other works use various models to reduce the environment and interaction to a textual space; perceptions are described in text by naming objects and features in the environment. An LLM then selects which object to choose, and other models perform robotic movement and action~\cite{dorbala2023can,wu2023embodied}. This can be used to respond to direct human instructions, or to anticipate human needs from implicit queries~\cite{kakodkar2023cartier}. Oftentimes instructions from humans are incidental and unstructured and it is important to be able to understand and act upon these~\cite{tang2023saytap}, or else an ambiguity exists which can only be resolved by considering the current situation alongside the human's instruction~\cite{zhi2024pragmatic}. Plans such as this can also be generated from a combination of the task description and the current state of the robot~\cite{ha2023scaling}. It is possible to ``freeze'' the LLM; place it into the architecture and train other models around it without altering the LLM's parameters. Smaller models can then be used to adapt what the frozen LLM outputs as a plan such that it can better be understood and performed by the robot~\cite{huang2022language}. An advantage of this approach is that it scales well. The process should theoretically be the same whether the lowest level tasks are atomic or are still quite high-level. Large environments can be operated in with large sets of tasks handled by a system of this type, with the planning agent solving numerous subtasks iteratively towards the initial goal~\cite{rana2023sayplan}. A single LLM can also be used to model a team of robots and divide work between them based on human input. For example, a high-level task can be broken down and assigned to robots based on their available skills~\cite{kannan2023smart}.

An interesting variant of this approach takes textual input and uses it to generate an imagined image of what the end goal should look like. Employing the DALL-E~\cite{ramesh2021zero} image generation model, this method follows a similar process, but by using an image as its immediate goal, it is potentially describing the task in a way closer to its perception. The concrete problem of properly understanding the task in text form is here moved slightly to a new form (as of course the generated image by the model may not align with the human's intention). A useful advantage of this approach however is that the robot's understanding of the goal is presented to the user at the start of execution, not the end. This may give the human more time to intervene in the likely event that the machine has misunderstood or hallucinated~\cite{kapelyukh2023dall}. Another means of utilising the LLM outside of the core control loop of the robot is to generate simulation data on which the robot can train. This is similar in that it fundamentally generates the conditions for the robot, and another model such Reinforcement Learning (RL), with the likely advantage of more explainability and predictability, can be used to actually perform the execution~\cite{wang2023gensim}. The same idea can also be used to generate alternative prompts and images for a more end-to-end approach~\cite{yu2023scaling}. Trajectories can also be learned and imagined from textual inputs and initial environment state, which are then used to define the inputs to a standard robotics transformer model~\cite{gu2023rt}.

\subsubsection{Writing Code and Defining Rewards}
Although a key advantage of using language is to replace the requirement for a user to understand and engage with technical command and code, a viable approach is to use the LLM as a translator. A human can define a behaviour in natural language, and the agent uses this to write code for the robot. Such code can be written to define a reward function for an RL agent. This utilises the twin advantages of LLMs for interpreting human commands, and RL for training the robot to achieve the behaviour~\cite{yu2023language}. This approach allows the LLM to both train the model and to incrementally update the reward function based on current performance~\cite{ma2023eureka} and, if available, visual feedback~\cite{narin2024evolutionary}. It can also help to build a policy that functions well, but can be steered by human input towards one of several acceptable options~\cite{hu2023language}. A similar approach is to write evaluation methods to label successful or unsuccessful trials~\cite{ha2023scaling}. Alternatively, an LLM can directly write code to manipulate the robot, just as a human would. By giving the model an API of functions with descriptions, it can write a new method to control the robot based on human-written comments and using a library of provided functions~\cite{singh2023progprompt,huang2023visual,liang2023code}, or can directly break down an objective and write the method with syntax guidance~\cite{hu2024deploying}. This kind of approach is sometimes augmented with other models to empower it, such as using Monte-Carlo Tree Search (MCTS) to find the best solution which is partially constructed and judged by a transformer model~\cite{zhang2023planning}. A key point of distinction between these approaches is how the environment and action affordances are presented to the robot. A popular solution is to provide example code in libraries, but fake methods can also be used, for example, the robot is given a method to detect objects which outputs a set of interactable elements in front of it; by presenting the information to the system in this manner, it naturally flows into the coding environment and allows the robot to consider it in the same way it would any variable from its corpus of coding examples. High-level abstractions like this also allow pretrained or manually written skills to be treated as a callable method without a requirement for the robot to engage with the specifics of how this is performed~\cite{lin2023text2motion}. The input space in such examples can be produced with segmented maps of objects in the environment, and other inputs such as gestures~\cite{lin2023gesture}.% Similar to direct code is the option of providing a library of nodes along with a command and having the LLM generate a behaviour tree~\cite{lykov2023llm}. 

\subsubsection{Direct Integration of Human Input}
LLMs or language embeddings can be inserted directly into the model in a manner where a human could not feasibly replace the LLM. Instead of using the language in a logical, sequential way as with the other categories, this approach is more ``black-box'' like; placing some text or a textual embedding into part of the model or its input, and training other models around this. Fundamentally, this still means that the language is used to instruct and command the robot, but not in an explainable or interpretable way. Some works place the LLM directly into the model for the robot, such as by using a combination of image labelling and extracted keywords from a human command to recognise landmarks during navigation~\cite{shah2023lm}. This sometimes demands a more careful model which considers that human input is sometimes incorrect, and this should be recognised and adapted to wherever possible~\cite{taioli2024mind}. LLMs frequently form part of the action selection process, such as mapping human instructions to the most similar action generated from environmental observations~\cite{yow2024extract}. Other approaches use a single LLM to process the instruction alongside representations of the environment perception or robot state~\cite{ni2023grid,zhu2024incoro}. Human input can also be fed into an LLM to adapt the inputs to another model~\cite{wen2024object}, even another, frozen, LLM~\cite{barmann2023incremental}. Approaches of this form facilitate an updating control loop if required, where the model is updated and trained based on success, failure, or human alterations and corrections~\cite{liu2023llm}. This type of structure could be promising in a lifelong learning context where robots can learn from their mistakes and correct undesired behaviours or shortcomings in the future. The LLM itself can even be called upon to generate better prompts for its own model to train on after failure~\cite{bing2023meta}. Already functional behaviours can also be subtly modified with human alterations by appending a representation of a command relative to an object of interest, such as instructing the robot to ``keep a much bigger distance from the computer'' when delivering drinks to a human~\cite{bucker2023latte}. Similarly, a policy can be trained that alters actions given by a human on a handheld controller, combining this control with a textual instruction and outputting joint parameters that satisfy both~\cite{karamcheti2022lila}. These approaches allow a human to subtly improve the performance of an already trained and functional model without a requirement for further training or alteration~\cite{shi2024yell}. This method of altering robot behaviour may be highly practical for non-technical users to work alongside robots as it may not be practical to finetune a robot to a specific task or environment; instead having the users advise it to subtly change using language could be much more feasible in future industry~\cite{cui2023no,huang2023voxposer}.
Another approach is to use behavioural cloning to learn from humans tele-operating the robot while narrating their tasks. This allows the policy to factor a linguistic instruction into the input space of the model during training, so that the policy can be used to control the robot with spoken commands during execution~\cite{lynch2023interactive}.

\subsection{Robot to Human Communication}\label{sec:R2H}
Agents are often difficult to understand for human operators and bystanders~\cite{rosenfeld2019explainability}, and a sensible application of LLMs is to address this problem. Robots can use language to describe their actions, beliefs, and intentions to a human; such an approach removes the requirement for expert users who can debug the robot's beliefs, instead communicating in a way that almost all can understand --- language. The generative aspect of LLMs is naturally suited to this task, as well as the question-answering mode that is often employed~\cite{wu2017visual}. This is especially relevant when paired with visual models and/or in safety-critical scenarios; describing why an agent took a certain action could help operators catch mistakes when human lives are at risk. The ``Robot to Human Communication'' category describes work which employs LLMs to feed information back to humans for purposes ranging from explainability to concisely representing their observations of the environment.

\begin{table}[ht!]
\centering
\caption{A selection of works using language for Robot-to-Human feedback. The ``LLM'' column indicates the most recent LLM they used (not all if they used several variants). ``Application'' is left high level to allow papers to be further subgrouped.}
\label{tab:LitRev_R2H}
\begin{adjustbox}{max width=\textwidth}
\begin{tabular}{@{}clll@{}} % Adjust column specifiers (l, c, r) as needed
\toprule
Modality & Paper & LLM & Application \\ % Table headers
\midrule
\multirow{10}{*}{\rotatebox{90}{Explainability}}
& \citeTable{Rankin_2023} & LINGO-1 & Describing actions \\
& \citeTable{zhang2023explaining} & GPT-4 & Describing actions \& Question-Answering \\
& \citeTable{gonzalez2023using} & GPT-3.5 \& Alpaca & Describing behaviour \\
& \citeTable{chen2023driving} & GPT-3.5 & Describing actions \\
& \citeTable{liu2023reflect} & GPT-4 & Describing task \\
& \citeTable{lu2023closer} & GPT-3.5 & Question-Answering \\
& \citeTable{raman2013sorry} & Syntax trees & Describing feasibility \\
& \citeTable{gong2018behavior} & N/A (templates) & Signaling intentions \\
& \citeTable{garcia2018explain} & N/A (templates) & Question-Answering \\
& \citeTable{das2021explainable} & Syntax trees & Diagnosing errors\\
\midrule
\multirow{12}{*}{\rotatebox{90}{Asking for Help}} 
& \citeTable{ren2023robots} & PaLM-2L \& GPT-3.5 & Asking when certainty is low \\
& \citeTable{mullen2024towards} & GPT-4 & Asking when certainty is low \\
& \citeTable{hori2024interactively} & GPT-3.5 & Asking to resolve ambiguity \\
& \citeTable{padmakumar2022teach} & Episodic Transformer & Asking to resolve ambiguity \\
& \citeTable{dougan2022asking} & N/A & Asking to resolve ambiguity \\
& \citeTable{banerjee2021robotslang} & LSTM & Describing env. \& asking for guidance \\
& \citeTable{sattar2011towards} & N/A & Asking when certainty is low \\
& \citeTable{song2023llm} & GPT-3 & Asking to resolve ambiguity \\
& \citeTable{thomason2020jointly} & N/A & Asking to resolve ambiguity \\
& \citeTable{zhang2023sgp} & GPT-3.5 & Asking to guide through decision tree \\
& \citeTable{wang2024safe} & GPT-3.5 & Asking to resolve ambiguity \\
& \citeTable{zheng2022jarvis} & BART & Asking to resolve ambiguity \\
& \citeTable{liang2024introspective} & GPT-4 & Asking to resolve ambiguity \\
\bottomrule
\end{tabular}
\end{adjustbox}
\end{table}

\subsubsection{Explainability}
Typically, language models which are employed for robot-human communication will simply output messages one-way to the user. This may take the form of the agent logging its actions and decisions, and using language models to efficiently communicate the relevant parts of this data. Alternatively, question-answering models can be connected to the agent to provide the user with a direct interface to the system. The user can now query the system about its reasoning for certain actions or which parts of the environment it is currently paying the most attention to.

Early works (before LLMs were commonplace) used predefined structures such as syntax trees to define text-based outputs to the human~\cite{raman2013sorry}. This may take the form of hand-written labels based on common outputs such as failure messages~\cite{das2021explainable}, or templates which are adapted to generate explanations of behaviour~\cite{gong2018behavior}. Answers to common debugging questions could also be generated to aid the human in understanding~\cite{garcia2018explain}. The overall theme from this area, however, was that there was a need for better verbal, language-based cues and explanations~\cite{han2021need}.

Since the introduction of LLMs, much more work has appeared in this space. LLMs can be used to analyse robot log files and explain their behavior~\cite{gonzalez2023using,sobrin2024explaining}, or they can answer questions in terms of the agent's reward function~\cite{lu2023closer}. Alternatively, complex descriptions of the reasoning behind actions can be generated in a multi-agent setting~\cite{zhang2023explaining}. This can include multiple sources of data such as by captioning audiovisual information to describe the solving of a task~\cite{liu2023reflect}. A promising application of this concept is explaining the behaviour and decision-making of autonomous vehicles~\cite{chen2023driving}, either by narrating as the vehicle drives, or by using a question-answer prompt system to allow a conversation with the agent about what it is focussing on and how it intends to act~\cite{Rankin_2023}. The dialogical approach, whereby the user can actively ask the model questions, may be a more natural form of explainability that extends from a language-based instruction method~\cite{nwankwo2024conversation}.

\subsubsection{Asking for Human Assistance}
Into the robot-human category, we also place robots that use the human to support them on request. These systems generally plan their actions and output them to a human who can interrupt if a robot made a mistake. The key driver of this approach is deciding when to ask for the human to help; it is imprudent to flood a human operator with questions and requests because they become overloaded. Approaches of this type typically compute some kind of confidence score and use that to determine whether they should proceed~\cite{sattar2011towards,mullen2024towards}. The system could also describe the problem it is facing and ask for human instruction~\cite{banerjee2021robotslang,thomason2020jointly}. A more sophisticated approach is to determine the ambiguity which is causing the robot to be unsure. This is a natural behaviour in humans to clarify by asking a follow-up question. This can take the form of asking for additional details about the human's command, and the details of what they want to be achieved~\cite{hori2024interactively,zhang2023sgp,zheng2022jarvis}, or could be used to resolve simple issues that the robot is struggling with such as location or task status~\cite{padmakumar2022teach}. Robots which resolve ambiguities by asking a question which rephrases the task or specifically targets the area of difficulty (for example asking ``is the object you want to the left of X?'' outperform those which ask for a simple rewording of the same task~\cite{dougan2022asking}. A well-designed approach of this type will highlight the issue to the user, or will provide a possible point of reference such as ``I cannot find X, but I can see Y'' so that the user can confirm how it should proceed~\cite{song2023llm}. This has implications on safety; robots can ask a human for assistance if they are not sure what the safest acceptable course of action might be~\cite{wang2024safe}. Finally, robots can frame their question as more targeted than a \textit{yes} or \textit{no} answer. In a scenario where ambiguity is present, the robot can choose those optional actions with the highest confidence and ask for clarification between them directly to the user~\cite{ren2023robots}, or else present some candidate actions from which the human selects~\cite{liang2024introspective}. This allows the human to more quickly resolve the exact issue with a specific command instead of giving a yes or no answer that might devolve into further issues. 

\subsection{Robot to Robot Communication}\label{sec:R2R}
Multi-robot systems often have a communication component, where robots exchange bids, plans, observations, or any other information. Classically this has been a carefully structured process based on predefined protocols. LLMs, however, provide the opportunity to remove these constraints and allow generative models to produce messages in a dialogical manner. From this dialogue emerges a collective intelligence which supports coordination, planning, and knowledge transfer. This approach is particularly appealing as it models a core strength of human groups using the exact same tools; language allows teams of people to organise, and the same may be true for robots. We term this type of system ``Robot to Robot Communication''.

\begin{table}[ht!]
\centering
\caption{A selection of works using language for Robot-to-Robot communication. The ``LLM'' column indicates the most recent LLM they used (not all if they used several variants). Application is listed, as all use language in the same way, but the use cases are often different. The number of agents and whether a Human-in-the-Loop is possible (HITL) are also included.}
\label{tab:LitRev_R2R}
\begin{adjustbox}{max width=\textwidth}
\begin{tabular}{@{}clllll@{}} % Adjust column specifiers (l, c, r) as needed
\toprule
Modality & Paper & LLM & Application & \#Agents & Human in loop \\ % Table headers
\midrule
\multirow{9}{*}{\rotatebox{90}{Role Playing}}
& \citeTable{li2023camel} & LLaMa-7B & Reasoning & 2 & \checkmark \\
& \citeTable{chan2023chateval} & GPT-4 & Debating & 3+ &  \\
& \citeTable{hong2023metagpt} & GPT-4 & Developing Software & 2+ &  \\
& \citeTable{wang2023adapting} & GPT-4 & Finding objects in \textit{ALFWorld} & 2 &  \\
& \citeTable{fu2023improving} & GPT-4 \& Claude & Bartering (buyer/seller) & 2 & \\
& \citeTable{nascimento2023self} & GPT-4 & Bartering (buyer/seller) & 2+ & \\
& \citeTable{zhang2023exploring} & GPT-3.5 & Reasoning & 2+ &  \\
& \citeTable{zeng2022socratic} & Various & Captioning & 3 & \checkmark\tablefootnote{The authors include people as a means of input to their language model component} \\
& \citeTable{wu2024shall} & GPT-4 & Planning & 2+ & \\
\midrule
\multirow{6}{*}{\rotatebox{90}{\parbox{2.3cm}{Inter-Agent Coordination}}}
& \citeTable{park2023generative} & GPT-3.5 & Modelling a community & 2+ &  \\
& \citeTable{bigazzi2023embodied} & CLIP & Navigating and captioning & 2+ &  \\
& \citeTable{zhang2023building} & GPT-4 & Fetching items in a home & 2 & \checkmark \\
& \citeTable{wu2023autogen} & GPT-4 & Reasoning & 2+ & \checkmark \\
& \citeTable{chen2023agentverse} & GPT-4 & Reasoning & 1+ &  \\
& \citeTable{papangelis2019collaborative} & N/A & Planning through dialogue & 2 &  \\
& \citeTable{du2023improving} & N/A & Reasoning & 2 &  \\
\midrule 
\multirow{3}{*}{\rotatebox{90}{\parbox{0.8cm}{Inter-Robot}}} 
& \citeTable{kannan2023smart} & GPT-4 & Planning and allocating tasks & 2+ &  \\
& \citeTable{mandi2023roco} & GPT-4 & Planning and coordinating & 2+ &  \\
& \citeTable{hunt2024conversational} & GPT-4 & Planning and coordinating & 2+ & \checkmark \\
\bottomrule
\end{tabular}
\end{adjustbox}
\end{table}

\subsubsection{Role Playing}
Language models such as GPT generally have a ``system message''. This is a prompt that the model considers with each message it sends; it can be thought of as the agent's ``ego''~\cite{PromptEngineering}. For example, in assistant-style systems like ChatGPT, this is often set to something like ``you are a helpful assistant''~\cite{chi5dialogue}. In practise these can be very long and complex and are often not fully published, but overall the system prompt defines the high-level behaviour and themes of the LLM. This can also be used to define a character for the agent, such as a tendency to be more direct, humorous, or polite. More formally, recognised personality types can be modelled and reflected by the LLM's outputs~\cite{safdari2023personality}. This practice is called ``role playing'' and allows a human to interact with a specialised agent, or can be used to create a dialogue between two chatbots, each of whom acts as though talking to a human, whilst assuming their defined character or style~\cite{li2023camel}. An approach of this form can allow an LLM to critique itself for iterative improvement~\cite{ye2023selfee}, or can use dialogue to encourage one agent to instruct another step-by-step on how to solve a task~\cite{wang2023adapting}. This can be expanded beyond two agents to a team, each of whom is given a slightly different role to play~\cite{chan2023chateval}. This mimics the way human teams are often constructed with a diverse combination of skillsets, backgrounds, and personality types~\cite{pieterse2006software,pollock2009investigating}, which is often suggested to be beneficial in certain settings~\cite{higgs2005influence,meslec2015balanced}. Although typically decentralised and peer-to-peer, such teams can also be modelled after hierarchical structures. A team of agents each with a different, role-playing specialism can be built, and as each works on a group solution they alter it and pass it on to their subordinate. For example an engineer can write code and pass it to a QA tester who validates its effectiveness, and so on~\cite{hong2023metagpt,qian2023communicative}. This allows different specialisms to be combined, although it is worth considering that the system prompts do not alter an agent's actual knowledge or skills. Future systems may finetune different models to this end. These processes often require substantial human guidance, although a teacher-student system can be used to automate some of this process~\cite{rasal2024llm}. Similarly, feedback and criticism from experience can be directly used to update such agents, allowing a group to criticise each other and collectively improve~\cite{fu2023improving}, or else to compete with each other and each attempt to get the best outcome from a collective exchange~\cite{nascimento2023self}. The process of forming optimal teams of differently configured language agents is being actively explored and may become a crucial part of coalition formation in future~\cite{zhang2023exploring}. Finally, multiple ways of interpreting the environment can be combined and treated as separate parts of the total model. This allows other types of data such as images and audio to be directly factored into the collective model. Dialogue can be an effective interface between these models as they are each fundamentally text-orientated~\cite{zeng2022socratic}.

\subsubsection{Inter-Agent Coordination}
Most closely matching an ad-hoc human group, inter-agent coordination describes groups of agents who share a common central repository such as a message board. Any agent can view this board, consider the current messages, and write a reply which is posted. This concept arises from ChatGPT, which models an agent conversing with a human. By restructuring the conversation so that each agent perceives other agents as aspects of its human interlocutor, the conversation can function naturally, with each agent participating and considering the suggestions of others. This has a key advantage of being feasible to organise with heterogeneous agents and no prior setup (provided a chat-board can be maintained, the only constraint is that each agent can read and post). Additionally, conversation forms the ideal basis for the merging and sharing of diverse knowledge and experience. Agents may have experienced different things and have built up a different understanding of their environment. It is not always practical to merge models, exchange maps, or upload images. In conversation, however, agents can condense this information into language which encapsulates much of that knowledge, and insert it into the group consciousness.

A foundational work of this form places a set of agents in a simulated environment where they can move, act, and engage in dynamic dialogues with other agents. This forms a community in which each agent acts according to its identity and experiences; knowledge such as current events and gossip diffuses throughout the group. Although this work has no direct utility, it acts as a robust proof-of-concept for language-based agent collectives~\cite{park2023generative}. One challenge for this approach is deciding when to speak; language models are slow to run, and conversation requires efficiency. Agents should not constantly flood the network with messages and should instead determine whether they have something worth saying, otherwise listening to others. This can be achieved with a separate, lower-weight model which determines if the main language-generating model should be invoked~\cite{bigazzi2023embodied}. Another consideration from a work that does not employ LLMs is to direct messages to other agents as questions, prompting them to update the sender with what they know~\cite{papangelis2019collaborative}. As language is a dense storage of meaning, it can be useful in scenarios where decentralised agents have a high cost for communication. A single sentence can infer a large amount of knowledge about the environment, actions, and intentions to other agents~\cite{zhang2023building}. Such approaches can be extended to allow agents to internally reason and predict their neighbours' actions and intentions. This can be done with direct language exchange, or can be used to consider one's neighbour using an LLM without explicitly communicating with them; essentially using language to facilitate robot-to-robot coordination and teamwork~\cite{zhang2023proagent}. Different architectures can be explored for inter-agent coordination, such as a hierarchy with a supervisor agent or a decentralised joint chat. Such architectures also facilitate the placement of a human in any position, essentially extending this approach to encompass the human-robot and robot-human categories in the same system~\cite{wu2023autogen}. The same fundamental approach can be used for the entire structure of the team while it pursues its goal. A task can be provided to agents who recruit the best team of experts, decide on their strategy, and enact the plan; language is the basis of each step~\cite{chen2023agentverse}. The core intuition here is that a team of communicating agents can reach a better solution through debate than a single agent can from thinking alone~\cite{du2023improving}.

\subsubsection{Inter-Robot Coordination}
The coordination of robots using language is not a well-explored area, but a few works do exist. The most clear example is RoCo, which sets a team of robot arms simple tasks and has them coordinate using language. Each robot speaks to the others in a shared chat log to propose actions and approve each others' plans. It also responds to mistakes such as collisions~\cite{mandi2023roco}. Another work has two robots discuss their solution to a high-level task in a debate format, resulting in paths of rooms through an environment. This approach also allows human approval of proposed plans by calling the human once agents agree; the human can also intervene during execution~\cite{hunt2024conversational}.

\subsection{Robot Control and Reasoning}\label{sec:Ri}
Language and language-based models can also be used internally within a single robot. This can take a variety of forms, but in general, these approaches seek to utilise the intelligence of LLMs within the robot's own control and reasoning processes.

\begin{table}[ht!]
\centering
\caption{A selection of works using language for robot control and reasoning. The ``LLM'' column indicates the most recent LLM they used (not all if they used several variants). \textit{Italics} indicate a simulator, and ``Application'' is left high level to allow papers to be further subgrouped.}
\label{tab:LitRev_Ri}
\begin{adjustbox}{max width=\textwidth}
\begin{tabular}{@{}clllll@{}} % Adjust column specifiers (l, c, r) as needed
\toprule
Modality & Paper & LLM & Physical Robot / \textit{Simulation} & Application \\ % Table headers
\midrule
\multirow{5}{*}{\rotatebox{90}{\hspace{-0.4cm}\parbox{1.5cm}{Transformer Robotics}}}
& \citeTable{brohan2022rt1} & RT-1 & 7-DoF Robot Arm & Describing task \\
& \citeTable{brohan2023rt} & RT-2 & 7-DoF Robot Arm & Describing task \\
& \citeTable{driess2023palm} & PALM-E & Various & Describing and planning \\
& \citeTable{padalkar2023open} & RT-X & 7-DoF Robot Arm & Describing task \\
& \citeTable{ahn2024autort} & AutoRT & 7-DoF Robot Arm & Describing task \& alignment \\
\midrule
\multirow{11}{*}{\rotatebox{90}{Language Architecture}}
& \citeTable{wang2023prompt} & GPT-4 & \textit{Various} & Acting on joints \\
& \citeTable{huang2022inner} & InstructGPT \& CLIPort & UR5e Arm & Planning and acting \\
& \citeTable{ulmer2024bootstrapping} & OpenLlama & N/A & Guiding through behaviours \\
& \citeTable{lykov2023llm2} & Alpaca 7B & N/A & Building behaviour trees \\
& \citeTable{cao2023robot} & GPT-3.5 & N/A & Building behaviour trees\\
& \citeTable{wei2022chain} & GPT-4 \& PaLM & N/A & Planning and reasoning \\
& \citeTable{yao2024tree} & GPT-4 & N/A & Planning and reasoning \\
& \citeTable{du2023video} & PaLM-E & Various & Planning and acting \\
& \citeTable{besta2023graph} & GPT-3.5 & N/A & Planning and reasoning \\
& \citeTable{xie2023reasoning} & GPT-4 & Unitree Go1 & Planning and reasoning \\
& \citeTable{gu2023conceptgraphs} & GPT-4 & Clearpath Jackal & Planning and reasoning \\
\midrule
\multirow{6}{*}{\rotatebox{90}{\parbox{1.5cm}{Dynamic Adaptation}}}
& \citeTable{shirai2023vision} & GPT-4 & Robot Arm & Re-prompting from error messages \\
& \citeTable{shek2023lancar} & GPT-4 & \textit{spot-mini-mini} & Modifying terrain type \\
& \citeTable{skreta2023errors} & GPT-3 & Franka Panda & Reprompting from error messages \\
& \citeTable{jin2023alphablock} & GPT-4 & Franka Research 3 & Providing feedback \\
%& \citeTable{bhat2024grounding} & GPT-4 \& PALM-2 & \textit{VirtualHome} \& Franka Research 3 & High and Low-level planning \\
& \citeTable{bhat2024grounding} & GPT-4 \& PALM-2 & Franka Research 3 & Planning \\
& \citeTable{yao2023retroformer} & GPT-3 & N/A & Question-Answering \\
\bottomrule
\end{tabular}
\end{adjustbox}
\end{table}

\subsubsection{Transformer Robotics}
Language models are typically built on the transformer architecture, which is defined by an attention mechanism and connections between the encoder and decoder. Although not exclusive to language, generally they are used to map language inputs to language outputs~\cite{vaswani2017attention}. A similar architecture can, however, be modified by making the output space a language-based representation of robot parameters~\cite{brohan2022rt1}. Combining this with a Vision-Language Model (VLM) creates an architecture which takes in an image of the environment and a textual task, and maps this in one step to a robot action which can be executed in real life~\cite{brohan2023rt}. This can generalise to different tasks using the same policy, and can also generalise to different robots, essentially transferring learning across platforms and tasks~\cite{padalkar2023open}. Further to this, language-based policies can be combined with pre-written skills and human teleoperation, each of which exists as part of a total architecture for training and execution~\cite{ahn2024autort}. These approaches offer an opportunity for language to bridge the gap between multiple domains to form a generalised, platform-agnostic, multimodal model for robotic systems~\cite{driess2023palm}.

\subsubsection{Language-based Architectures}
Some architectures use language internally or to facilitate other means of robotic control. This can be highly effective in bringing together different component models and combining them through a language interface. An example is InnerMonologue~\cite{huang2022inner}, which uses language to create an internal conversation between different aspects of the robot. The robot might call the scene descriptor to announce what it can see, after which the robot checks with the human what they want, and then the robot executes an action and asks the success detector if it has achieved its goal. Each part of the architecture is ``talking'' to each other within the overall structure. Such ``self-talk'' is a powerful tool and may be useful in a number of domains including navigation~\cite{xie2023reasoning}. Other approaches use language as the entire policy by providing textual descriptions of the environment and actions, and use the LLM to generate joint parameters in one step~\cite{wang2023prompt}. More broadly, some systems create a complex architecture with each component being language-based. From environment representation to action, language is the representation space~\cite{wang2024describe}. Another way of doing this is to pre-generate responses to your expectations and use the LLM to best fit the situation to those responses. For example, a dialogue tree can be created and the LLM is asked to match a user's input to whichever response most closely matches the expected answers on the tree~\cite{ulmer2024bootstrapping}. Alternatively, the most likely user inputs can be predicted and responses can be pre-generated; this is something that could feasibly done for robotics as most user requests are likely to be predictable in the context of the overall mission or goal~\cite{chi5dialogue}. Some approaches use LLMs in conjunction with another representation of information such as using graphs to perform basic mathematics~\cite{cao2023enhancing}, or representing knowledge in a graph structure to be queried and reasoned through as a tree~\cite{li2024enhanced}.

LLMs can also be employed to process only the abstract and high-level parts of the problem, for example by deciding where to go or guessing where an object is likely to be, with all other actions performed using more standard approaches~\cite{gu2023conceptgraphs}. An alternative method is to use language to generate a policy that is then used. The classic method of doing this is to generate a behaviour tree which then controls the robot~\cite{cao2023robot,lykov2023llm}. The behaviour tree approach can also be used to control a team of robots~\cite{lykov2023llm2}. Being next-symbol predictors, LLMs sometimes struggle to reason about questions, especially when careful consideration of several factors is involved. A proposed solution is to encourage them to give their reasoning as they answer --- a ``chain-of-thought'' approach~\cite{wei2022chain}. An extension of this approach is to create a ``tree-of-thoughts'' which considers multiple paths and includes backtracking to find the best reasoning~\cite{yao2024tree}. An improvement on this tree structure is a graph structure which also supports backtracking and aggregation of thoughts into the final plan~\cite{besta2023graph}. This can further be developed by combining visual information; by breaking down the task into a tree structure and employing text-to-video models, a complete plan can then be generated including both task specifications and video demonstrations extrapolated from the initial state and future actions~\cite{du2023video}. The trend of equipping LLMs with an understanding of other data structures is common across multiple works and appears to be a core means of empowering language-based systems~\cite{wang2024chain}.

These complex decision-making problems, especially within an environment which is difficult to interpret and understand make video games a natural test case for these approaches. Language has been applied as a planning tool in \textit{Minecraft}~\cite{zhu2023ghost}, and as an instruction space for planning and action for agents whose learning can transfer across multiple games~\cite{AbiRaad2024Scaling}. Each of these developments increases the complexity of language-based environmental reasoning that can be performed, a problem that is discussed later in section \ref{sec:Lim:env}.

\subsubsection{Dynamic Adaptation with Language}
Some systems use language as a feedback mechanism with which to adjust an existing model. For example, a functional policy for a robot to walk can take in a textual description of the environment and can use this to modify the behaviour accordingly~\cite{shek2023lancar}. This provides a promising means for robots to adjust to unseen conditions through the generality of language models. By invoking the LLM to check each step of a plan, iterative adaptation can be achieved within the model's architecture. This can include checking each step of the plan one-by-one~\cite{raman2022planning}, or calling on external observations from other sources to update, improve, or complete each step of the plan~\cite{xu2023rewoo}. Textual feedback can also be used to catch mistakes when a robot's plan is converted to a Planning Domain Definition Language (PDDL) and the plan is symbolically checked; error messages can be passed back to the model for the plan to be adjusted accordingly~\cite{shirai2023vision}. This same approach can also be used for parsing error messages from robot execution, which in turn are used to edit generated code in the same way a human might respond to error messages while programming~\cite{skreta2023errors}. Alternatively, simple assessments of why a plan did not work are then converted into summaries and revised plans~\cite{shinn2024reflexion}. Generally speaking, these methods use the environment (whether that be a physical environment or a human) as a means to evaluate their solution and then iteratively improve the output through this process~\cite{yao2023retroformer,wu2023plan}. Another approach is to generate a language plan step by step but to execute the plan as the steps are generated. After a step is completed, an environment update is fed back into the model so it receives real-time feedback and can adjust accordingly~\cite{jin2023alphablock}. Adaptation can also be used to adapt a shared plan iteratively for different goals or to resolve different issues, for example, a ``brain LLM'' can write a high-level plan which is adjusted by a ``body LLM'' to make it performable by the robot. This allows a single plan to be written, improved, and ultimately realised by successive language calls~\cite{bhat2024grounding}.

\section{Applications}\label{sec:App}
There are many applications of language-based robots, but we have provided four primary use cases which cover some of the key opportunities and challenges.

\subsection{Autonomous Vehicles}
Although human drivers do not generally have the ability to communicate verbally with each other, some amount of implicit communication is employed (pointing, flashing lights, etc). It is possible that the integration of LLMs will provide agents an interface with which to perform the complex coordination required on road environments with a wide variety of vehicles each with different needs and each pursuing different end goals. The changeable nature of road systems such as contraflows, breakdowns, and weather is very difficult to train for with existing approaches, so LLMs may provide a useful means to interpret and handle unexpected scenarios~\cite{cui2024survey}. A few works are actively pursuing LLMs as a control method for vehicles, taking in textual descriptions and sometimes images, and outputting a text-based command for how the steering wheel and pedals should be adjusted~\cite{cui2023drivellm,cui2024drive,xu2023drivegpt4}. A key advantage of using LLMs in this way is that they can provide reasoning and justification, either as an observational tool to explain the actions of an existing model~\cite{park2024vlaad}, or to pair a reason with the predicted action of the human, which in turn could be used to control the vehicle~\cite{wen2023dilu}. The poor ability to reason about the environment, along with obvious safety concerns likely render a fully integrated language-based self-driving car unlikely. Some have addressed this with hybrid approaches, combining the LLM with a more reliable rule-based planner~\cite{sharan2023llm}. However, the most practical avenue for LLMs in this domain is language as a tool for explanation (robot-to-human), control at a very high level (human-to-robot), and inter-agent negotiation e.g. at a stop junction (robot-to-robot). Such negotiation is naturally suited to LLMs, whether in a competitive setting~\cite{wu2024shall}, or a more cooperative system~\cite{hunt2024conversational}. The explainability problem is especially well-matched to autonomous vehicles, and is being actively explored~\cite{tekkesinoglu2024advancing}; language is an especially promising avenue in this space and naturally extends the way humans might explain their intentions and actions while driving~\cite{yuan2024rag}.

\subsection{Language-based Explainability of Robot Behaviour}
Explainability for language models is a popular area of research with a wide variety of sub-areas; the literature is still young and LLMs are still fundamentally a black-box technology that is difficult to understand~\cite{zhao2024explainability}. Despite this, language is a natural domain for explanation, especially for non-technical users. LLMs are themselves used to explain the behaviour of other models~\cite{gonzalez2023using}, offering the additional advantage of open dialogue where a human can ask specific targetted questions of the model~\cite{lakkaraju2022rethinking}. Robotics has an especially high demand for explainability, as robots tend to be mistrusted in general~\cite{lewis2018role}, and can often cause damage or loss of life if they malfunction~\cite{setchi2020explainable}. The union of language and robotics creates an area with an absolute need for explainability and transparency in the absence of trustworthy execution. Robots that use language would be able to explain their failures~\cite{das2021explainable}, or could continuously update operators and surrounding humans of their reasoning, intentions, and beliefs. This type of explainability may prove crucial to providing safe and trustworthy AI systems in the future.

\subsection{Ad-hoc coordination}
As robots and other autonomous systems become more prevalent it will be necessary for teams of such devices to dynamically form and cooperate~\cite{stone2010ad}. For example, several home robots may need to organise a group strategy to clean a home, a team of agricultural robots may need to combine efforts to expedite a task before the weather turns, or a set of helper robots might even need to collaborate to save a human in danger. Each of these requires a coordination and control structure that is agnostic to the models and morphology of the platforms involved~\cite{flavin2024bayesian}. Language is already used by humans in this way and is an important aspect of effective ad-hoc teams~\cite{white2018facilitators,mirsky2020penny}. Collaborative emergent behaviour can be achieved through the use of language~\cite{li2023theory}. LLM-based structures may also provide a means for differences in robot ability to be considered and optimised for during allocation and planning~\cite{han2024llm}. In these types of systems, a variety of possible topologies may be required featuring peer-to-peer or hierarchical structures, each requiring an organisational system which could benefit from language as a communication medium in the same way humans operate~\cite{guo2024large}. In order to establish a shared understanding, robots may need to agree upon names and labels for objects in the environment based on sensor observations, after which point language can be used to represent the complex details of this specific entity when coordinating towards a goal~\cite{ekila2024decentralised}. Finally, this type of coordination naturally extends to human-robot teams~\cite{liu2023llm,gong2023mindagent}.

\subsection{Cobots}
The popular emerging model for robots in the workplace is the Collaborative Robot (``cobot'') model. This means that humans are not replaced with machines, but instead, humans and robots work alongside each other. Many governments are pursuing this model to improve their local industries, especially for small-to-medium-sized businesses without the need or budget for large and specialised equipment~\cite{lefranc2022impact}. A key advantage of this is that systems and processes could feasibly be structured to utilise the best of each party's skills and abilities. Much research and many robot platforms have been applied to this problem~\cite{taesi2023cobot}. Use cases of real deployment have been performed with reasonable success~\cite{el2018working}. Many different approaches have been proposed ranging from working simultaneously on the same widget to forming a human-robot production line with sequential work~\cite{el2019cobot}. Regardless, this means of interacting with robots will likely soon be commonplace in not just the factory setting, but in the service and medical domains too~\cite{djuric2016framework}. Substantial trust issues exist due to the dangerous nature of industrial robots~\cite{kopp2022linguistic} which may be abated somewhat through the careful use of language. For humans to effectively work alongside robots there must be a shared understanding and means of communication, at least in one direction as a way to task the robot (human-to-robot)~\cite{bergman2019human}, and language is likely the most powerful and natural approach~\cite{ionescu2021programming}. The inclusion of language in these robots' policies in any of the ways described in this work could substantially resolve trust issues and provide an effective means for humans and robots to cooperate in this setting.

\section{Challenges}\label{sec:Lim}
There are still many challenges associated with Language Models, both generally and in robotics. In this section, we discuss some of the primary areas of concern.

\subsection{Core Issues with LLMs}
LLMs have several core problems which introduce issues in any autonomous system to which they are applied. These are not specific to robotics, but should still be considered as they present obstacles to the development of language-based robots.

\subsubsection{Hallucination}
The term ``hallucination'' refers to foundational models inventing or imagining new information that typically has little to do with the inputs or internal knowledge. This problem is well documented and has persisted despite strong concerns that this may render the models almost unusable~\cite{rawte2023survey}. It has been suggested that due to the fundamental architecture and approach of these models, hallucination is impossible to fully eliminate~\cite{xu2024hallucination}. The attention has turned to mitigating this problem, primarily through detection and resolution, although approaches are varied~\cite{tonmoy2024comprehensive}. The exact reasons behind hallucination are still not well understood, although some have attempted to analyse and find traces of the problem by investigating hidden states in the model~\cite{duan2024llms}. Some have proposed calling in additional knowledge to resolve the issue~\cite{liang2024learning}, or considering the likelihood of the generated text (for example in vision-language models, a bird near a tree is a sensible output so is more trustworthy)~\cite{zhou2023analyzing}. Another approach which is relevant to the robot-to-robot category is to collaborate with other agents; some agents can then abstain from a group decision if required~\cite{feng2024don}. The most popular area of development is to detect hallucination so the outputs of the model can at least be ignored~\cite{chen2023hallucination,leiser2024hill,zhu2024knowagent}. This is essential as, like the closely related adversarial examples in image recognition~\cite{wiyatno2019adversarial}, hallucination events generally give highly erroneous results without a correspondingly low confidence; the model does not know it is ignorant. This has been partially addressed by pairing the outputs of the model with an associated uncertainty value~\cite{kirchhof2024pretrained,han2024towards}. 

Hallucination is an issue in all LLM applications, although in robotics it may have more impactful negative consequences. An imagined danger which does not exist or an imagined empty space where there is in fact a human could result in a robot causing considerable harm or loss of life. Evidently for an embodied robot to be deployed in the real world, issues surrounding hallucination must be addressed. At the absolute least, such events should reliably be detected so the robot does not misclassify a human, or attempt to act upon an environment it does not truly understand.

\subsubsection{Security}
Any control system for a robot can be subjected to cyber-attacks. Hostile actors may seek to gain control of or manipulate the robot in order to disrupt industrial processes or ransom a company by interrupting its operations. Connecting an LLM to one of these systems makes it a target. This is especially true with the increasing ubiquity of LLMs directly interacting with code; an area with substantial potential but also a lot of risk~\cite{yang2024robustness,yao2024survey}. These models are being deployed in other areas of security already, including writing test cases~\cite{zhang2023well} and modelling assertions~\cite{kande2024security}. Despite their potential utility in defending other systems, the LLMs themselves are highly vulnerable; the vast majority of currently available language models are affected by a new family of attacks, and despite these being specific to LLMs, the defenses are limited~\cite{liu2023prompt}.  The closely related field of adversarial machine learning provides many examples of slight manipulations in a model's input causing substantial changes in its output~\cite{wiyatno2019adversarial}. This holds true for generative models in a similar way, where semantically identical but subtly reworded prompts generate incorrect outputs for robotic tasks~\cite{wu2024safety}.

Due to fears of LLMs being used for malicious intent, they generally have ``guardrails'' integrated~\cite{dong2024building}. The exact details of these are not made public, but it is likely that the system message has certain prompts which forbid them to answer questions which are deemed unacceptable. The attack known as ``jailbreaking'' circumvents these guardrails through intelligent prompting~\cite{greshake2023not}. It is difficult to define what constitutes unacceptable inputs to an LLM, and these are often defined by crowd-sourced judgement~\cite{thoppilan2022lamda}. This is closely related to injection attacks which convince the LLM to output privileged information or even run code by accessing a website~\cite{wu2024new,liu2023prompt}. Defenses are currently limited and mostly respond to a specific set of test data with little guarantee that further modification of the attack payload will not defeat the defenses~\cite{chen2024struq}. Some approaches employ a conversational approach to finding and preventing security issues, either through conversation with a human~\cite{xu2021bot}, or between agents~\cite{zeng2024autodefense}. LLMs are often closed-source black box models with unknown architecture. Specific attacks can be used to discover and effectively steal some fraction of the LLM at low cost, which could offer attackers the ability to reverse-engineer and compromise the system~\cite{carlini2024stealing}. Other attacks include appending a random set of characters to the end of the prompt, which causes the LLM to answer a forbidden question~\cite{cao2023defending}. Some methods check whether the LLM should respond before it is called, catching the problem early~\cite{ruan2023identifying}, whereas others generate the response and check that this is safe to do before taking action. The black-box nature of LLM guardrails can make them difficult to counter, although a promising technique entices the model to explain why it rejects a query, and then uses its own language to invoke a forbidden response~\cite{deng2023jailbreaker}. LLMs can also be fooled by asking leading questions that hint at the wrong answer~\cite{mei2023assert}. There is an inherent trade-off between safety and helpfulness that must be carefully tuned depending on the application~\cite{zhang2023defending}. 

Regardless of the details, a vulnerable LLM which is connected to a robot or multi-robot system poses a considerable threat if bystanders could (deliberately or accidentally) trick it into performing unsafe or undesired actions. For language-based robots to be effectively deployed, there is a considerable need for stronger guarantees of security.

\subsubsection{Training Cost}
Large foundational models have a very high training cost. This becomes particularly problematic if the environment is non-stationary (the reward or expected state of the environment changes over time)~\cite{de2023llm}. More generally, if a lot of training effort is applied to a large model, it is specific to the current environment and goals much the same as in biological organisms~\cite{dawkins2017}. When the situation changes, which is very likely in robotics, the model may need to be adapted. This issue is not specific to LLMs, and LLMs may have the ability to adapt to the changes more intelligently than RL and other approaches, as their basis of training on diverse text gives them some experience of the implications of novel scenarios~\cite{shek2023lancar}. Future robotic systems may need to be more flexible to changing demands, and so modular architectures may need to be employed to facilitate this~\cite{liang2022transformer,devin2017learning}. Regardless, without new methods to resolve this, LLMs are likely to be expensive and time-consuming to train.

\subsection{Issues with LLMs in Robotic Systems}
Some issues apply more specifically to the application of LLMs to robotics. These are not unique to robotics, but are less likely to be a focus of the main LLM community, so should be considered separately from those in the previous section.

\subsubsection{Modelling the Environment}\label{sec:Lim:env}
Modelling the world has been a crucial problem in robotics for decades~\cite{brooks1991intelligence}. Being essentially next-word prediction engines (``n-gram'' models)~\cite{brown1992class}, LLMs inherently do not include a representation of the environment and it can not easily be extended into their architecture. It has been argued that generalisation itself can only be effectively achieved through an understanding of the world which allows an agent to solve a new task in context~\cite{mitchell2023ai,mitchell2023comparing}. Understanding and planning at a human level will likely require the ability to reason hierarchically on multiple levels of abstraction and on different timescales~\cite{lecun2022path}. While they can often prove effective in games and tasks where there is an imagined world, they tend to prove lackluster when asked to model an environment and track plans into the future. Examples such as stacking blocks are very difficult for LLMs to effectively solve~\cite{valmeekam2024planbench}. Much work in this space aims to provide LLMs with a means of representing the environment that can be manipulated through text, the theory being that this would allow them to effectively understand the world and their effects upon it; a requirement that most deem important for an embodied robot when planning into the future~\cite{jordanous2020intelligence}. At the very least, it is necessary for a coherent language-based plan which can be understood and approved by humans.

A first-principles approach can be employed to this end which tests hypotheses about the environment directly in a physics engine~\cite{liu2022mind}. This is likely too fundamental to be practical. More promising approaches maintain a textual description of the world state which is predicted and then updated by the robot's actual actions~\cite{yoneda2023statler}. Alternatively, a simulation can be used as a world model~\cite{xiang2024language}. The most common approach is to define a PDDL which the LLM can use to describe and update the environment~\cite{silver2022pddl,silver2023generalized}. The key intuition here is that formal logic is text-based but provides a concrete and reliable way to model and understand the state of the world. Such an approach can effectively model relationships between objects and agents~\cite{liu2023llm,guan2024leveraging}. This can also be achieved with templated queries asking the LLM to describe within clearly defined expectations some relationship between objects. For example, ``X is next to Y, how far should A be from B?''~\cite{ding2023task}.

An environment can also be imagined as an image; image generation models such as DALL-E can be used to create an image of what the goal state should be, and this is used to guide the model~\cite{kapelyukh2023dall}. Otherwise, LLMs can be used to resolve ambiguities in how objects might feasibly exist and be placed within an environment~\cite{zhang2024rail}. This style of interpreting the environment visually may require more advanced models which can comprehend the nature of the physical world; in the same way, transformers are suited to reasoning with language, similar architectures may prove useful in the domain of video~\cite{selva2023video}, along with rich embeddings of video data~\cite{sun2019videobert}. With recent developments in generative video models, it is possible that such architectures are on the horizon~\cite{videoworldsimulators2024}. More complex relationships and environment understanding can also be modelled using representations like trees and graphs~\cite{rana2023sayplan}. Formalising a model of the environment in this way is sensible as such models are clear and reliable, the challenge is how the LLM queries and updates it; for such solutions to be practical there is a need for strong handling of the interface between them~\cite{chen2023not}. One approach to this is to build two separate models, one classical LLM and another grounded model which interprets the environment. Their decisions can be combined in the same approximate means as \textit{SayCan}~\cite{brohan2023can}, to perform the action which best satisfies both the overall plan, and the environmental state~\cite{huang2024grounded}.

Although simple tasks may be tackled by robots without a good understanding of the environment, the field of language-based robotics is likely to be constrained by this insufficiency. Future models and technologies which tackle this problem have the potential to revolutionise the way robots plan, act, and operate.

\subsubsection{Memory and Nonstationarity}
In dealing with both non-stationary, changing tasks and user demands, and the potential damage and wear on the robot itself, some degree of adaptation and lifelong learning may be required~\cite{liu2021lifelong}. This may consist of the agent adapting its policy, learning from its mistakes to become better at complex tasks, or even tuning its understanding of how to operate alongside humans~\cite{irfan2021lifelong}. Doing this would require a robot to update itself in one of two ways: (1) by finetuning its model constantly throughout its life, or (2) by storing a history or set of beliefs which are updated through experience. Finetuning may have issues around overfitting to the specific everyday tasks the robot is required to do (for example it may forget safety protocols which are not often needed). It also requires storing the complete LLM, unless a technique such as Low-Rank Adaptation (LORA) is used~\cite{hu2021lora}. The option of storing history may be more promising, with LLMs emulating memory~\cite{park2023generative,barmann2023incremental}. A wide variety of techniques are actively being explored, and are well described by \citeTable{wang2024survey}. At a high level though, three key issues arise: (1) what information to store, (2) how to store and summarise it, and (3) how to access and use those memories. Storing relevant information in the short term is already achieved with the standard context window, although eventually, memories will pass out of this ``short-term memory'' and will require a ``long-term memory'' system alongside this~\cite{ng2023simplyretrieve,yao2023retroformer,zhang2024large}. It can still be difficult for the LLM to pick out what is useful in this way~\cite{kuratov2024search}. An extension of this idea is to summarise interactions by extracting the key sentiment and storing these as memories~\cite{park2023generative}, or to use embeddings or vector representations to capture some essence of the memory~\cite{zhong2024memorybank}. Such memories can be of an event, entity, or even a skill~\cite{wang2023voyager}. Once a database of memories has been collated, access can be achieved through Retrieval-Augmented Generation (RAG) instead of feeding it into the main context window~\cite{gao2023retrieval}. Some degree of compression and forgetting may be required to maintain a useful record which is not bloated with information, much as humans do~\cite{hou2024my}, or else careful consideration should be given of which information the LLM has already internalised to minimise unnecessary access or saving of information~\cite{huang2024learn}.

For robots to be useful in complex and changing environments, they will need to be able to remember and call upon knowledge, skills, and experiences from a long time prior. To do so they will need a solid means of storing, condensing, and recalling information in a way that allows them to iteratively learn and adapt to real-world environments and tasks.

\subsubsection{Uncertainty of LLMs and Integrating them with other Models}
LLMs are inherently probabilistic, and so introduce uncertainty into their outputs~\cite{adam2023large}, this is known as \textit{aleatoric} uncertainty --- the output of the model varies each time it is run due to a variation in the selected next token. This is not uncommon in machine learning models, however, the sequential nature of language (or a language-like series of actions) means that the small uncertainties compound and result in outputs increasingly divergent from what may be expected from the initial inputs~\cite{ling2024uncertainty}. In a more practical sense, slight variations or unpredictabilities in a plan at the nascent stage are then built upon and a robot's actions at a later point might be drastically different from what the designers intended. It is also important to consider \textit{epistemic} uncertainty --- There can also be situations in which the knowledge of LLMs or their conception of the environment is incorrect~\cite{felicioni2024importance}. In this case, it is more feasible to argue that such deficiencies can be corrected through other models or checking with humans. Some approaches use the LLM itself to evaluate its uncertainty, although this can be argued as moving the problem to another unreliable model~\cite{tanneru2024quantifying}. Other approaches incorporate more reliable models alongside the LLM to give an estimate of confidence, such as conformal prediction~\cite{wang2023conformal,wang2024safe}. This could potentially be used to reduce the level of uncertainty in the system and improve the predictability of a language-based robot.

Regardless of the means of reducing the uncertainty, this presents a significant problem for LLMs in robotics. It is likely that the most effective way of harnessing the opportunities of LLMs is to use them for planning at a high-level, and potentially for handling unexpected inputs, after which the LLM is used in conjunction with other models, or as a component of a complex architecture. Relying on RL and other models for the end-level actions is likely to be the most practical~\cite{shi2023unleashing,hu2023enabling}, or else using LLMs to empower the training of RL~\cite{chen2024vision} or create a step-by-step plan to be solved with RL~\cite{shukla2023lgts,dalal2024plan}. Language can act as the high-level operator of a set of APIs and other models which can be called upon like ``tools''~\cite{shen2024hugginggpt,qin2023toolllm,schick2024toolformer}.

\subsubsection{Computation Speed and Memory Size}
LLMs are inherently slow to generate responses. Although parallelism is often employed during training it is not practical to distribute next-word prediction across different devices~\cite{brakel2024model}. Many LLM-based robotics approaches operate at reasonably low frequencies (1-10Hz)~\cite{brohan2023rt}, whereas practical robots are likely to require a frequency much higher (in the region of 500Hz)~\cite{chignoli2021humanoid}; it is not practical to run the LLM at anything close to this frequency for computational power reasons. Further to this, LLMs generally require a lot of memory to store, and are often not available to the public except through an API, which incurs additional latency. It is unlikely that language could practically be used for split-second decision-making, where simpler reactive policies are likely to be more reliable. Further to this, an API-based approach requires a constant online connection --- something which is also impractical in many use cases, especially those where robots may be operating in hazardous environments with poor or damaged infrastructure.

\subsubsection{Lack of Robot Datasets}
Datasets are often a limitation for robotics as gathering data can be expensive and time-consuming. LLMs may solve this issue, both through the virtue of training on a broad set of data which is generally freely available online~\cite{liu2024datasets}, and by generating their own training data using the LLM itself. This can take the form of self-generated instructions to broaden the range of training data based on a few human-generated inputs~\cite{wang2022self}. Alternatively, human responses can be used to identify bad generations and used to expand the dataset on which to train~\cite{hancock2019learning}, or to reinforce the intended role or expected style of the agent~\cite{bae2022building}. Human labelling can also be used to identify if an image represents a success in a textually-defined task, which may be useful in training models to do this~\cite{nair2022learning}. Despite this, there are currently limitations on the amount of data and the number of simulation options available to developers of language-based robotics in certain cases~\cite{liu2024libero}, and more mature options would be useful in future.

\subsection{Ethical and Safety Concerns}
The deployment of LLMs in robotic systems presents several issues of an ethical nature; for robots to be reliably integrated with LLMs these should be addressed.

\subsubsection{Safety}
AI safety as a concept predates most modern AI technologies, having been discussed at length in science fiction at least as early as Asimov's famous three laws of robotics~\cite{asimov1941three}. While these laws neither function particularly well in his writings, nor are practical to implement in most cases, their general theme persists in much of the AI safety literature. It should be noted, however, that much of this research is not responding to an Armageddon-like threat~\cite{mccauley2007ai}, rather subtle issues such as bias and fairness, or lack of alignment with human morals when lives are at stake~\cite{leslie2019understanding}. Some have employed inter-agent debate with human judgement as a strategy to deduce human moral codes~\cite{irving2018ai}. Generally, however, many concrete problems exist which are difficult to overcome~\cite{amodei2016concrete}. Language models need only emulate human morals in the domain of text, which in some senses may be easier than doing so in action.

The problem of aligning LLMs to human morals is an active area of research~\cite{wang2023aligning}. Firstly, defining human morals is incredibly difficult. Some have resorted to language-based constitutions, which are derived from Asimov's laws themselves~\cite{ahn2024autort}. Others have attempted to codify morals with Linear Temporal Logic~\cite{yang2023plug}. This can either be injected into the system prompt before generation of text~\cite{hua2024trustagent} or afterwards as a form of post-plan inspection~\cite{yuan2024r} which can then be used to reject or alter the output~\cite{liu2022second}. As morals are so difficult to define, many have turned to humans to provide them. Using a crowd-sourced labelling effort, models can be trained to inherently understand morals without a set of rules to follow~\cite{ji2024beavertails}, although this is likely to be less trustworthy in practice. One other aspect of safe and trustworthy language models is the issue of multilingual interface; LLMs generally operate best in the English language as that is what predominates their training data. Many are trained to have some ability in other languages but this tends to be limited~\cite{kassner2021multilingual,guo2020wiki}, and safety measures are often less effective on ``low-resource'' languages that the model has less training data for~\cite{shen2024language}. It would be prudent to ensure that those who interact in other languages have a similar experience and ability in order to ensure fair and trustworthy use by a wide variety of people~\cite{chen2023phoenix}.

Some efforts have sought to use datasets of safe and unsafe inputs and outputs to modify existing models such that they align better with human ethics. One method is to use the LLM itself to generate this data and further train itself~\cite{xu2023baize,wang2022self}, although the scope of this is unlikely to be reliable in all cases. Alternatively, humans can be inserted into a training loop for an LLM~\cite{ouyang2022training}. In real-world use cases, guarantees of safety are unlikely to be strong, and when language-based robots inevitably make mistakes then the most important outcome is that they can be learned from. Given an existing model, just a few safety examples can be used to finetune it~\cite{yang2023shadow}, although a careful balance must be pursued to ensure the model is not ``paralysed'' by safety~\cite{bianchi2023safety}. A model can also be pruned to reduce its size~\cite{sun2023simple} which retains most of the power and can increase safety~\cite{hasan2024pruning}, although the consequential reduction in ability may make this impractical. 
Simple approaches to safety in language-based robotics limit the parameters of the output such as maximum acceleration~\cite{yoshikawa2023large}, some define instinct-level behaviour to protect the robot and surrounding people~\cite{zhang2023bridging} while others incorporate a complex set of handwritten rules governing the entire output~\cite{ahn2024autort}. Regardless of the efficacy of safety approaches, trust will likely always be an issue with any autonomous system; people may be reluctant to rely upon a system they do not understand unless they are given appropriate evidence and experience that it is reliable~\cite{zhou2024don,tekkesinoglu2024advancing}.

For language-based robots to be trustworthy, a combination of handwritten rules and learning from mistakes will likely have to be employed. There is a substantial issue with public trust in either case, and research which provides reliable guarantees of safety would be highly impactful. Regardless, safety concerns are likely always to be relevant in the context of operating a large or unconstrained robot, especially in the proximity of humans.

\subsubsection{Trust, Responsibility, and Governance}
A complete study of trust in LLMs is beyond the scope of this work, for a more in-depth discussion we refer the reader to the work of \citeTable{sun2024trustllm}. The architecture of communication in LLM-based systems we describe goes beyond the simple ``dyadic'' (1 human and 1 agent) style of interaction that is often studied and instead introduces a complex relationship of trust between different parts of the total collective~\cite{schneiders2022non}. In these circumstances, the specification of trust becomes more challenging and relies on a variety of fields and ideas~\cite{abeywickrama2023specifying}. A core issue in all aspects of autonomous system deployment, especially with LLMs is that their apparent ability can engender a very high level of trust from a user; while this may seem positive, it can cause people to rely on the system without checking its outputs. Mitigations for this problem do exist but are so far limited to rephrasing the agent's outputs slightly~\cite{kim2024m}. There is also a concern around the ``deskilling'' of operators of these systems who rely on effective systems and gradually lose their own abilities to verify the accuracy of the outputs~\cite{choudhury2024large}. Some have proposed that the outputs of an LLM should be empirically verified instead of checked by a human, which would avoid the over-trust issue, although this is not always possible, especially in robotics~\cite{zhou2024don}.

Ideally, a model should be able to match its output with an appropriate level of confidence. A model which can do this reliably is said to be well \textit{calibrated}; calibration in LLMs is hard to quantify but is generally quite poor~\cite{spiess2024quality}. This is especially true for LLM-generated code which is a popular option in the robotics domain, making trust calibration an important problem~\cite{virk2024enhancing}. The increasing prevalence of LLM-generated data online also raises the concern of a ``self-referential learning loop'' where agents learn from other AI-generated content~\cite{choudhury2024large}. The implications of this are yet to be seen but may include a loss of context or availability bias in the training data of future systems~\cite{Knapton_2023}.

There are also many ethical and legal issues surrounding the liberal and irresponsible use of LLMs, especially in domains such as healthcare where human lives are by definition always at stake~\cite{ong2024ethical,mesko2023imperative}. This is even now becoming an issue in the legal domain, with an apparent need for practitioners to make themselves aware of the limitations and pitfalls of these models~\cite{supreme2023}. There is an additional concern about who is responsible for the inevitable mistakes and potential harms caused by this technology; in a robotics context this could cause severe loss of life and it is not always apparent whether the responsibility lies with the designer of the robot, the LLM, or the deployer of the whole system into a working environment~\cite{yazdanpanah2023reasoning}. With LLMs becoming renowned for their abilities there is an apparent asymmetry between credit for positive outcomes and blame for negative ones~\cite{porsdam2023generative}.

As an emerging technology, LLMs face substantial obstacles in terms of gaining an appropriate and balanced level of trust from the public, forming part of a sensible allocation of responsibility, and being well-regulated and understood by the various stakeholders involved. When integrated into robots, this potentially becomes even more important, as robots can be dangerous when operating alongside humans. For language-based robots to be feasible there is a need for significant progress in the understanding of these issues.

\section{Conclusion}
As Language Models continue to improve and are integrated into a wider variety of computer systems, it is necessary to consider their utility and impact in the domain of robotics. Language can act as an input space for models which, through breakdown or generation of code and rewards, conveniently allow untrained human operators to command robots in the same way they would with human colleagues. This has ramifications for collaborative robotics and may facilitate the increasing prevalence and practicality of autonomous systems in people’s home and work environments. By the same logic, robots can use language as a means to communicate their actions and intentions to humans, as well as actively using language as a bidirectional interface for clarifications. This allows the twin advantages of humans and robots to be combined in an efficient way, again without the need for the human to learn a programming language or other technical details.

Extending this, language can be used as the communication space for groups of agents to coordinate and organise. Much like in human groups, language is generally a common means of communication that allows differently trained and aligned platforms to disseminate information as well as build complex collective plans. Finally, language and language-orientated models can be used as the fundamental basis for the architecture of powerful models. This permits an overall information pipeline between multiple humans and multiple robots where any or every component part could feasibly operate entirely on language, allowing human participation and oversight across the entire observation, decision, communication, and action process.

Language models are inherently limited in several ways. They suffer from hallucinations and lack a true understanding or model of the environment in which they operate. Additionally, both safety and security make robotics applications ethically and financially risky. These areas are actively being investigated and some approaches are promising. It is likely that language can never practically be used for the entire control flow of a robotic system. Rather, select parts of the whole can be language-based, with more reliable and tested approaches being used where required. Language is a fundamentally irreplaceable ingredient of human intelligence, both social and individual, making it an exciting and necessary avenue for investigation in robotics and likely a cornerstone of many future systems.

\begin{acks}
This work was supported by the UK Research and Innovation Centre for Doctoral Training in Machine Intelligence for Nano-electronic Devices and Systems [EP/S024298/1]. Figures in this work contain images from Flaticon.com
\end{acks}

\bibliographystyle{apalike}
\bibliography{main}
\end{document}